\documentclass[conference]{IEEEtran}
\IEEEoverridecommandlockouts

\usepackage{cite}
\usepackage{amsmath,amssymb,amsfonts}
\usepackage{graphicx}
\usepackage{textcomp}
\usepackage{xcolor}
\usepackage{booktabs}
\usepackage[T1]{fontenc}
\usepackage{lmodern}
\usepackage{microtype}
\usepackage[hidelinks]{hyperref}
\usepackage{capt-of}

\def\BibTeX{{\rm B\kern-.05em{\sc i\kern-.025em b}\kern-.08em
    T\kern-.1667em\lower.7ex\hbox{E}\kern-.125emX}}

\begin{document}

\title{Procedural Content Metageneration via Program Search and Continual Abstraction Discovery}

\author{
\IEEEauthorblockN{Matthew Siper}
\IEEEauthorblockA{Game Innovation Lab \\
New York University\\
New York, USA \\
ms12010@nyu.edu}
\and
\IEEEauthorblockN{Ahmed Khalifa}
\IEEEauthorblockA{Institute of Digital Games \\
University of Malta\\
Msida, Malta \\
ahmed@akhalifa.com}
\and
\IEEEauthorblockN{Julian Togelius}
\IEEEauthorblockA{Game Innovation Lab \\
New York University\\
New York, USA \\
julian@togelius.com}
}

% \IEEEoverridecommandlockouts
% \IEEEpubid{\makebox[\columnwidth]{979-8-3315-9476-3/26/\$31.00 \copyright2026 IEEE\hfill}\hspace{\columnsep}\makebox[\columnwidth]{ }}
\IEEEpubid{\makebox[\columnwidth]{979-8-3315-9476-3/26/\$31.00
\copyright2026 IEEE\hfill}\hspace{\columnsep}\makebox[\columnwidth]{ }}

% Full-width first-page teaser
\IEEEaftertitletext{%
    \vspace{-0.75em}%
    \begin{center}%
    \begin{minipage}{\textwidth}%
        \centering
        \includegraphics[width=\textwidth]
        {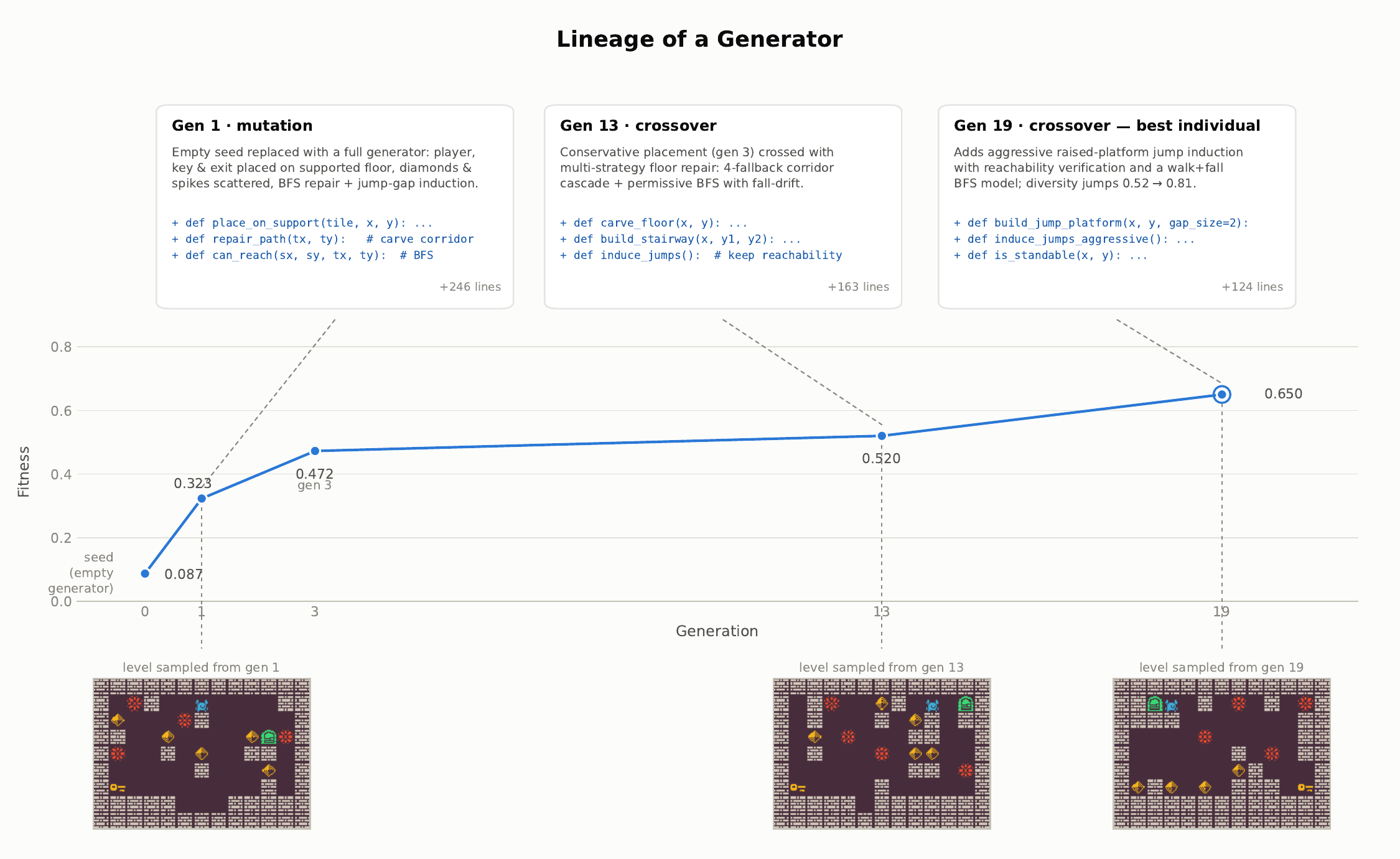}

        \captionof{figure}{
        Evolutionary lineage of the best Dangerous Dave generator. Each callout summarizes the code introduced at that step, and each image shows a level sampled from the corresponding generator. The new generators just need to be better than the starting seed to be accepted in the archive.
        }
        \label{fig:first-page-lineage}
    \end{minipage}%
    \end{center}%
    \vspace{-0.25em}%
}

\maketitle
\IEEEpubidadjcol

\maketitle
\IEEEpubidadjcol

%
% 1) First page out the best method's levels
% 2) Explain what the primitives actually are
% 3) Pick a generator's lineage to track and then underneath it show the level generated underneath it
% 4) 
%

\begin{abstract}
Large language models can generate executable programs, which makes it possible to search directly over procedural content generators rather than individual levels. We study this approach in Sokoban, Zelda, Dangerous Dave, and Lode Runner. Each run evolves complete Python generators through language-model mutation and crossover. We introduce Continual Abstraction Discovery, or CAD, which extracts reusable primitives from high-fitness programs into a run-specific helper module. A $2\times2$ experiment crosses CAD with access to a fixed hand-written domain API. The completed data set contains 160 complete runs, with at least ten 50-generation runs in every cell. CAD raises mean final best fitness in all eight domain and API comparisons. Across all CAD runs, learned libraries are adopted by most later programs and repeatedly rediscover validation, reachability, and structural utilities. These results support that discovering reusable primitives improves evolutionary program search for content generators.
\end{abstract}

\begin{IEEEkeywords}
Large Language Models, Procedural Content Generation, Level Generation, Program Synthesis, Code Generation, Genetic Programming, Evolutionary Search
\end{IEEEkeywords}

\section{Introduction}

Games that feature procedurally generated content usually rely on custom generators tailored to the mechanics, constraints, and design goals of a particular game. Creating these generators requires substantial engineering and repeated evaluation. Procedural content generator generation seeks to automate this process by searching for the generator itself rather than directly searching for individual artifacts~\cite{kerssemakers2012procedural}. This process can also be called procedural content \emph{metageneration}.

Existing approaches learn generators through evolutionary computation, machine learning, or reinforcement learning~\cite{kerssemakers2012procedural,khalifa2020multi,summerville2018procedural,khalifa2020pcgrl,liu2021deep}. Many produce neural or otherwise opaque representations. Such generators can be effective, but they are difficult to inspect and edit. Executable source code offers a different representation. A program can be compiled, tested, modified by a designer, and reused outside the search system.

Large language models are increasingly capable of writing and revising code. This makes it possible to place a code-generating model inside an evolutionary loop. A candidate generator is executed, its levels are evaluated, and the resulting feedback guides later mutation and crossover calls. Related systems have used language models as variation operators for executable program search~\cite{novikov2025alphaevolve,aygun2025ai,siper2025profit}. Procedural content metageneration presents a demanding instance of this problem since the quality of a generator can only be judged through repeated execution and behavioral evaluation.

Search over raw code fixes the vocabulary available to variation. As programs grow, useful routines for reachability, entity normalization, repair, and structural construction may be independently recreated in many candidates. A hand-written helper API can expose reusable operations, although its vocabulary remains fixed before search begins. We introduce \emph{Continual Abstraction Discovery}, or CAD, as a way to let the set of executable primitives increase during a run. CAD extracts reusable utilities from high-fitness programs into a run-specific helper module, validates the module, and refactors programs to use the learned utilities when this can be done safely.

We evaluate the system on four domains from the PCG Benchmark~\cite{khalifa2025procedural}. The main study crosses a fixed domain helper API with CAD. This design separates access to a reusable vocabulary from the discovery of a vocabulary during search. 

We make four contributions.
\begin{itemize}
    \item We demonstrate LLM-driven evolutionary search over executable Python-level generators in Sokoban, Zelda, Dangerous Dave, and Lode Runner.
    \item We introduce CAD, a search-time mechanism that extracts, validates, and reuses helper functions discovered within high-fitness programs.
    \item We present a $2\times2$ study that separates CAD from access to a fixed expert API across 160 completed runs.
    \item We show positive CAD effects across the 4 game domains and provide an analysis of the learned-library growth, adoption, and primitives.
\end{itemize}

\section{Related Work}

A wide variety of computational methods have been applied to generating game content, in particular, game levels. These range from evolutionary computation~\cite{togelius2011search} and constraint satisfaction~\cite{smith2011answer} to supervised learning~\cite{summerville2018procedural} and reinforcement learning~\cite{khalifa2020pcgrl}. Published games typically rely on constructive generators~\cite{shaker2016procedural} specialized for the particular game, which limits reusability.

Recently, researchers have applied LLMs to generating game levels~\cite{todd2023level,sudhakaran2023mariogpt,hu2024game,gallotta2024large,sweetser2024large,nasir2024word2world,nasir2023practical,jiang2026agentic}. The results are mixed. In order to represent a game level in a way that an LLM can understand, it needs to be converted to a string somehow. Typically, this means representing the level as a sequence of rows, where each character represents a tile. Initial results showed that fine-tuning generators on levels from particular games improved generation capacity, and that more modern and recent LLMs outperform older and smaller ones on level generation (as on most other things)~\cite{todd2023level}.

However, current LLMs have limitations when generating game content, particularly spatial content such as levels. For example, LLMs tend to have problems ensuring that clear paths exist from point A to point B, and that the right number of items and characters of various types exist (such as exactly one player character, and the same number of doors as keys). Other pathologies of LLM-generated game levels seem to originate in weak spatial reasoning~\cite{todd2023level,gallotta2024large,nasir2024word2world}.

One of the major use cases of LLMs is code generation. Software engineering is currently being transformed by the emergence of LLMs that can produce working code from an informal specification. However, in many cases, you know what you want but not exactly how to get there. In those cases, you can create a fitness function measuring what you want, and then embed the code-generating LLM inside an evolutionary loop. This approach was popularized by AlphaEvolve~\cite{novikov2025alphaevolve}, which, among other things, discovered an improved matrix multiplication algorithm, and similar methods have also been applied to creating scientific data analysis solutions and financial trading policies~\cite{aygun2025ai,siper2025profit}.

Using LLMs to generate code inside an evolutionary loop can be seen as a form of genetic programming, which is the application of evolutionary computation to program synthesis~\cite{poli2008field,koza1990genetic}. From this perspective, the LLM is carrying out the mutation and/or crossover operations in genetic programming. An important concept in genetic programming is Automatically Defined Functions, where a set of callable functions is evolved in tandem with the main program, constituting a form of evolved abstraction~\cite{koza1993hierarchical}.

Continual abstraction discovery (CAD) is also related to recent work on agent skill libraries, where reusable code or behaviors are stored for later use, such as Voyager's skill library for Minecraft agents~\cite{wang2023voyager}. The key difference in our setting is that the library is not a record of solved tasks or manually exposed skills, but a continually updated set of helper functions extracted from high-performing generator programs during an ongoing evolutionary search. CAD, therefore, functions as a search-time representation mechanism. It changes the program primitives available to later mutation, crossover, and refactoring calls while the population is still being optimized.

Returning to PCG research, work on PCG through machine learning~\cite{summerville2018procedural}, including reinforcement learning~\cite{khalifa2020pcgrl}, generally learns generators, which are then used to generate game content. However, they mostly learn neural networks or other types of black box representations, which do not allow for easy human interpretation and editing. There are exceptions to this, where evolution or similar methods are used to learn generators expressed in a symbolic representation~\cite{kerssemakers2012procedural,khalifa2020multi}. The approach presented in this paper can be seen as procedural procedural content generator generation, or procedural content metageneration, using a search-based approach in generator space enhanced by LLM code writing. A key invention beyond the combination of tool and domain is the Continual Abstraction Discovery mechanism, which is similar to Automatically Defined Functions but using LLM-driven evolution.

\section{Method}

\subsection{Program Representation}

Each individual is a complete Python program defining \texttt{generate(context\_dict)}. The function receives the grid dimensions and an initial level and returns a tile grid. A program may contain local functions, constants, and layout routines, while the public interface remains fixed across the population. This representation allows every candidate to be compiled, executed, inspected, and evaluated with the same harness.

All runs begin from one minimal seed program. The seed contains tile constants and a blank or copied initial grid, with no generation logic. It is evaluated once and becomes the root of the archive. The archive retains every successfully evaluated individual.

\subsection{Evolutionary Search}

At each generation, one parent is sampled from the archive with fitness-proportional probability. Fitness values are floored at $10^{-6}$ before normalization. When the archive contains at least two individuals, crossover is selected with probability $0.25$. A second parent is then sampled with the same rule. Failed crossover attempts fall back to mutation. 
% All random draws use a deterministic random number generator keyed by configuration seed, run number, and generation.

Mutation and crossover are performed by an LLM. The prompt contains the parent program or programs, a domain description, the accumulated run memory, and the fixed generator contract. Mutation asks for a strong behavioral edit. Crossover asks the model to combine compatible mechanisms from both parents into one complete executable program.

Every candidate passes through a robustness pipeline. The program must compile and complete at least three of five smoke-test executions. A failure triggers an LLM correction call conditioned on the error. The mutation or crossover cycle is attempted up to four times. A generation that never produces a valid candidate is recorded with fitness zero. Parse failures, compile failures, smoke-test failures, and timeouts are logged separately.

After every generation, an LLM reflection call appends a structured entry to a per-run memory file. Each entry records the attempted change, the observed result, and global lessons for later search. The accumulated global lessons are supplied to subsequent mutation and crossover calls in place of raw per-parent evaluation feedback.

\subsection{Evaluation and Fitness}

Each candidate is executed 30 times. The sampled levels are evaluated with the quality, validity, and diversity functions supplied by the PCG Benchmark~\cite{khalifa2025procedural}. We compute the fraction of fully valid levels, the mean benchmark quality, and the mean pairwise diversity over all sampled level pairs. Fitness is defined as 
\begin{equation}
\label{eq:fitness}
F = \frac{1}{3}\left(V + Q + D\right),
\end{equation}
where $V$ is the fraction of levels whose quality is equal to $1.0$, $Q$ is mean level quality, and $D$ is mean pairwise diversity. All terms lie in $[0,1]$. Candidates that crash or fail validation receive fitness zero. 
% The primary endpoint is final best fitness, defined as the maximum fitness among all individuals discovered during a run.

\subsection{Continual Abstraction Discovery}\label{sec:cad}

% CAD gives each run an initially empty helper module. 
One of the contributions of this work is continual abstraction discovery. Starting at generation 8 and at five-generation intervals thereafter, the system identifies programs at or above the 75th percentile of fitness within the run. An LLM extraction call analyzes these programs and proposes reusable utility functions from that code. 
% These utility functions are saved in a library that all the evolved programs can use during evolution. 
% The helper module is maintained under a fixed capacity.
An extracted function must compile and pass a smoke test. The system allows up to two compile corrections. The source program is then refactored to call the new helpers. A refactor is accepted only when the refactored program matches the original program on a fixed-seed behavioral test (the generated levels has similar behavior characteristics). Accepted refactors are re-evaluated and their archive entries are updated. Refactoring is disabled when its success rate remains below $30\%$ after a ten-generation warmup.The extracted functions are recorded in a system library that can be used by programs during evolution. 
% The system records library size, helper adoption, helper calls, and the line count of the best program. For the abstraction audit, semantically equivalent functions are grouped under a shared label. Total Calls aggregates the observed uses of each grouped abstraction.

\section{Experimental Design}

\subsection{Domains}

We evaluate four tile-based domains from the PCG Benchmark:
\begin{itemize}
    \item \textbf{Sokoban:} is a 5x5 Japanese puzzle game where the player needs to push boxes to specific target locations.
    \item \textbf{Zelda:} is a 7x11 dungeon crawler inspired by dungeon rooms from the original The Legend of Zelda game. The goal of the level is to get a key and reach the exit without getting killed by monsters. 
    \item \textbf{Dangerous Dave:}is a 7x11 platformer game based on game with the same name. The goal of the game is for the player to get a key and reach the goal without getting killed by spikes.
    \item \textbf{Lode Runner:} is a 11x16 puzzle platformer game based on a game with the same name. The goal of the game is for the player to collect all the gold and avoid getting caught by the enemies. The puzzle element is that the player can't jump; they can only navigate the level by walking, digging through the floor, climbing ladders, and moving through ropes. 
\end{itemize}
For all these problems, we use the framework's quality and diversity outputs. These outputs mainly use an $A^*$ solver to compute quality and use some behavioral and visual characteristics to compute diversity. Each call to a generator is limited to two seconds.

% Table~\ref{tab:domains} lists the grid size used in each domain. Solvability is checked with an $A^*$ solver limited to 100 nodes and 500 milliseconds per level. Each call to a generator is limited to two seconds.

% \begin{table}[t]
% \centering
% \small
% \caption{PCG Benchmark domains used in the main study.}
% \label{tab:domains}
% \begin{tabular}{@{}lc@{}}
% \toprule
% \textbf{Domain} & \textbf{Grid} \\
% \midrule
% Sokoban & $5\times5$ \\
% Zelda & $7\times11$ \\
% Dangerous Dave & $7\times11$ \\
% Lode Runner & $11\times16$ \\
% \bottomrule
% \end{tabular}
% \end{table}

\subsection{Factorial Conditions}

The main study tests 4 experiments that test the usefulness of having helper functions. To assess this, we have two parameters that we are testing:
\begin{itemize}
    \item \textbf{CAD:} allowing the LLM to abstract programs and generate helper functions as explained in section~\ref{sec:cad}.
    \item \textbf{Expert API:} having a group of hand-designed helper functions that are commonly used in PCG. The library contains validated primitives for entity normalization, reachability, repair, and domain-specific structural operations. Examples include \texttt{connect\_floors\_with\_ladder} and \texttt{ensure\_one\_player}. The expert API tests whether providing helper functions will be enough or CAD is still needed.
\end{itemize}
Every experiment uses the same minimal seed, evolutionary search, reflection memory, correction pipeline, evaluation budget, and fitness function. We ran each experiment 10 times for 50 generations.

% \begin{table}[t]
% \centering
% \small
% \caption{Conditions in the $2\times2$ factorial study.}
% \label{tab:conditions}
% \begin{tabular}{@{}lcc@{}}
% \toprule
% \textbf{Condition} & \textbf{Fixed API} & \textbf{CAD} \\
% \midrule
% Base, no CAD & No & No \\
% Base, CAD & No & Yes \\
% API, no CAD & Yes & No \\
% API, CAD & Yes & Yes \\
% \bottomrule
% \end{tabular}
% \end{table}

\subsection{LLM and Execution Configuration}

All language-model calls use GLM 5.2. Mutations uses temperature $0.2$. Correction and refactoring use temperature $0.1$. The mean 50-generation run used approximately 9.5 million input tokens and 2.6 million output tokens. The mean cost was approximately \$25 and the mean wall-clock time was approximately 2.5 hours. Each run stores per-generation statistics, program sources, evaluation analyses, prompts, valid rendered levels, the evolution tree, reflection memory, helper-module snapshots for CAD conditions, and token and timing logs.

\section{Results}

\subsection{Cross-Domain Search Performance}

\begin{figure*}[t]
    \centering
    \includegraphics[width=\textwidth]{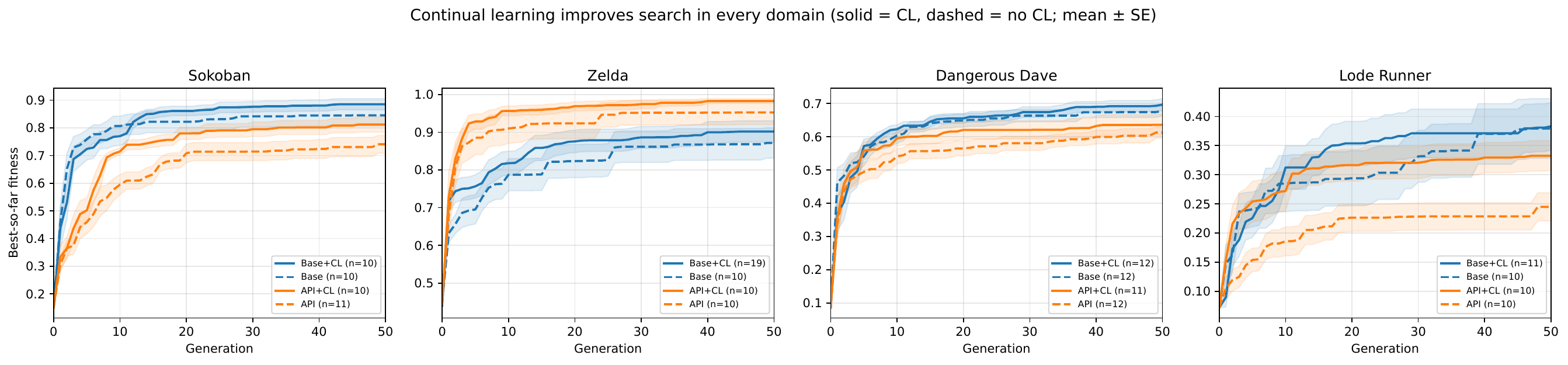}
    \caption{Best-so-far fitness over 50 generations in each domain. Solid lines show CAD and dashed lines show no CAD. Shading reports mean $\pm$ standard error, and legends show the completed run count for each cell. The plot uses CL as the implementation label for CAD.}
    \label{fig:domain-curves}
\end{figure*}

Figure~\ref{fig:domain-curves} shows a positive endpoint difference for CAD in every domain. The magnitude varies across games. The Base contrast is small in Lode Runner and larger in Sokoban. Under the fixed API, the largest separation occurs in Lode Runner, where the CAD trajectory continues improving while the no-CAD trajectory plateaus earlier. All experiments favor CAD, which gives an exact two-sided sign-test result of $p=0.008$. An interesting outcome is that only in Zelda, having a predefined API is better than having no API to start with. We believe that the simplicity of the problem benefited from the available functions, as the other problems need more complex solvers than just pathfinding and counting objects.
% The Base effects range from $+0.004$ in Lode Runner to $+0.040$ in Sokoban. The pooled run-weighted Base gain is $+0.049$ with $p=0.200$. The API effects range from $+0.023$ in Dangerous Dave to $+0.088$ in Lode Runner. The pooled API gain is $+0.051$ with $p=0.312$.

\begin{figure*}[t]
    \centering
    \includegraphics[width=\textwidth]{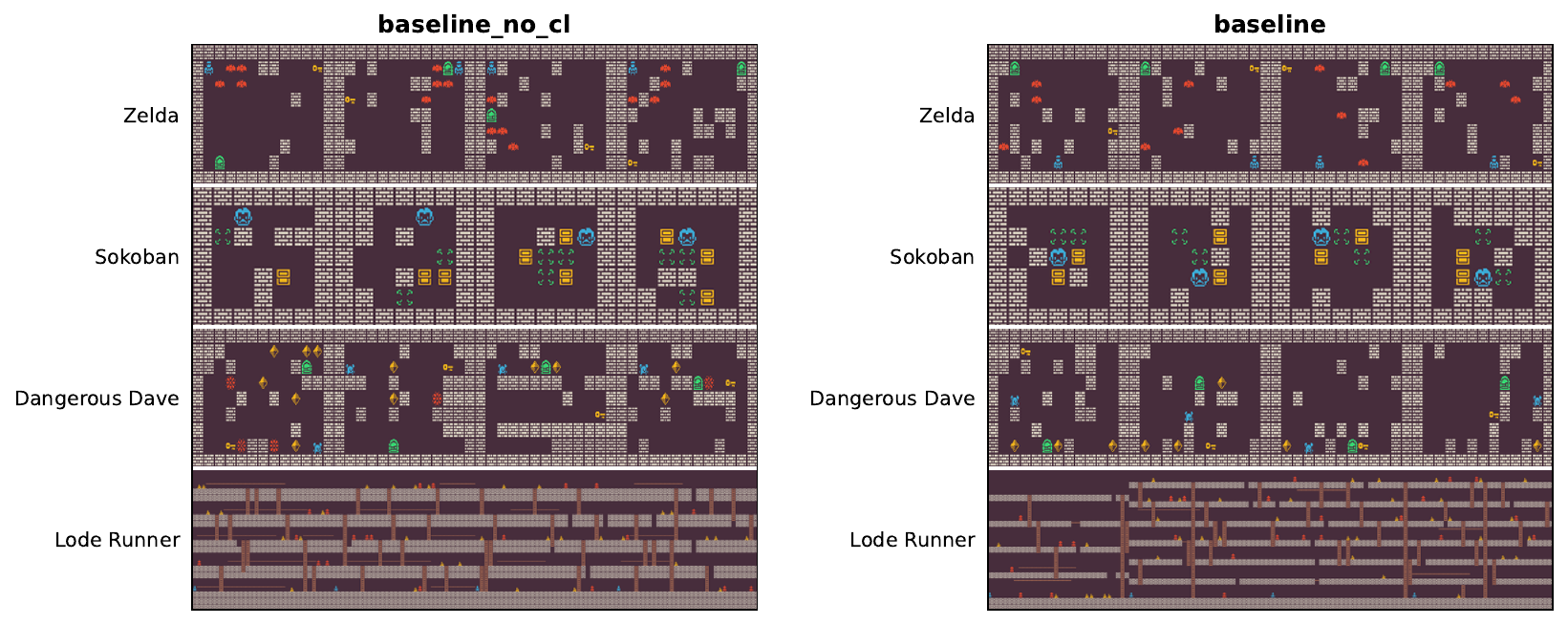}
    \caption{Representative generated levels from the Base condition without CAD on the left and with CAD on the right. Rows show Zelda, Sokoban, Dangerous Dave, and Lode Runner. Each row contains four samples. The images illustrate output structure and are not used as evidence of playability, which is measured through benchmark execution.}
    \label{fig:base-mosaics}
\end{figure*}

Figure~\ref{fig:base-mosaics} shows example of the generate levels for the best generator for every game. Looking at the generated levels, we can notice that the visual diversity of levels, especially in Dangerous Dave, is lower with CAD than without. We believe the reason behind that is that the diversity metric of Dangerous Dave only cares about the trajectory of the player solution and not the visual diversity, which makes these levels diverse even if they look similar.

\subsection{Learned-Library Growth and Adoption}

\begin{figure*}[t]
    \centering
    \includegraphics[width=\textwidth]{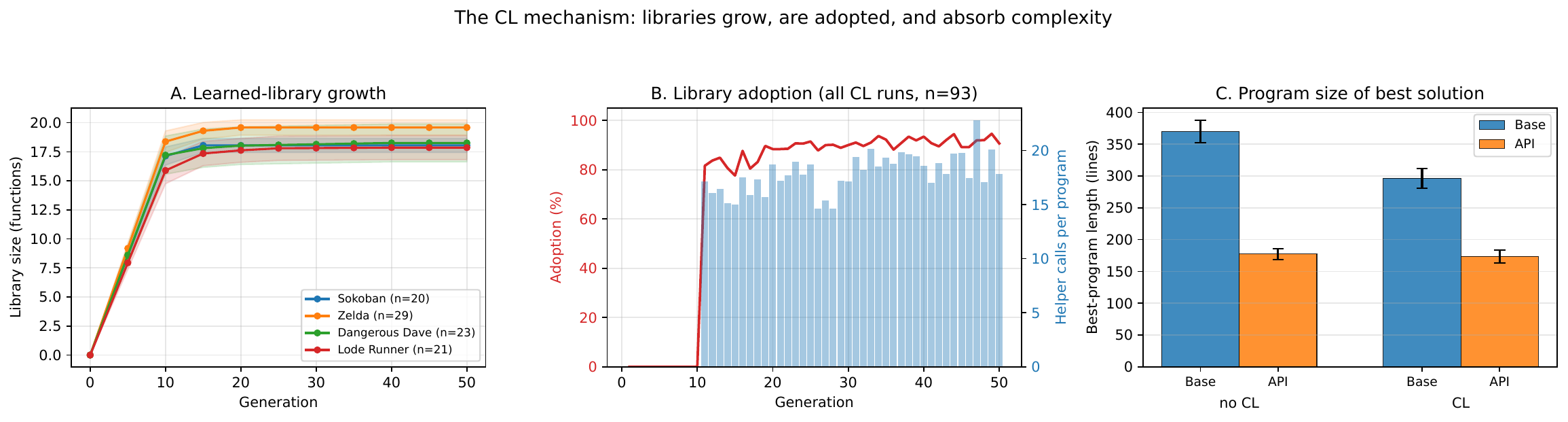}
    \caption{Mechanism measurements across CAD runs. Panel A shows learned-library growth by domain. Panel B shows the reported adoption rate and helper calls per program. Panel C compares the line count of the best solution under Base and API prompting with and without CAD. Shading and error bars report uncertainty around the mean. The plot uses CL as the implementation label for CAD.}
    \label{fig:cad-mechanism}
\end{figure*}

Figure~\ref{fig:cad-mechanism} shows that the learned libraries expand sharply during the early extraction cycles and then stabilize across all four domains. Once helper use begins, the reported adoption rate remains above roughly $80\%$ for most later generations, while programs contain approximately 15 to 20 helper calls on average.

The program-size comparison shows a second effect. In the Base condition, CAD reduces the mean length of the best program from roughly 370 lines to roughly 296 lines. Programs with the fixed API are already much shorter, at roughly 160 lines without CAD and 173 lines with CAD. This pattern is consistent with reusable logic moving out of the main generator program. The fixed API absorbs much of this complexity from the beginning, leaving less source-length reduction for CAD.

\subsection{Repeated Abstraction Discovery}

\begin{table}[t]
\centering
\small
\caption{Most frequently rediscovered learned abstractions across all CAD runs. Semantically equivalent functions are grouped, and Total Calls is the aggregate call count in the supplied audit. The complete audit is provided in the supplementary source.}
\label{tab:abstraction-freq}
% \resizebox{\columnwidth}{!}{%
\begin{tabular}{@{}llc@{}}
\toprule
\textbf{Abstraction} & \textbf{Category} & \textbf{Total Calls} \\
\midrule
In Bounds & Other & 4,693 \\
Entity Count Normalization & Validation & 3,848 \\
Ensure One Player & Validation & 2,611 \\
Grounded Empty Cells & Utility & 465 \\
Flood Fill & Other & 569 \\
Deep Copy Grid & Other & 118 \\
Find Player Position & Utility & 2,370 \\
Ladder Placement & Structural & 768 \\
Reachability BFS & Validation & 344 \\
Relocate Unreachable Entities & Validation & 320 \\
Connected Components & Other & 297 \\
Find Tiles & Other & 574 \\
\bottomrule
\end{tabular}%
% }
\end{table}

Table~\ref{tab:abstraction-freq} shows a recurring core of validation, reachability, and structural operations. \texttt{in\_bounds} and Entity Count Normalization each appear in 28 of the CAD runs. Ensure One Player and Grounded Empty Cells appear in 24 runs. Find Player Position, Ladder Placement, Reachability BFS, Relocate Unreachable Entities, and \texttt{connected\_components} each appear in 20 runs. CAD discovers both a shared set of broadly useful operations and a large tail of run-specific utilities. The repeated functions correspond to common generator requirements such as entity counts, traversal, connectivity, and tile placement.

\begin{figure*}[t]
    \centering
    \includegraphics[width=0.9\textwidth]{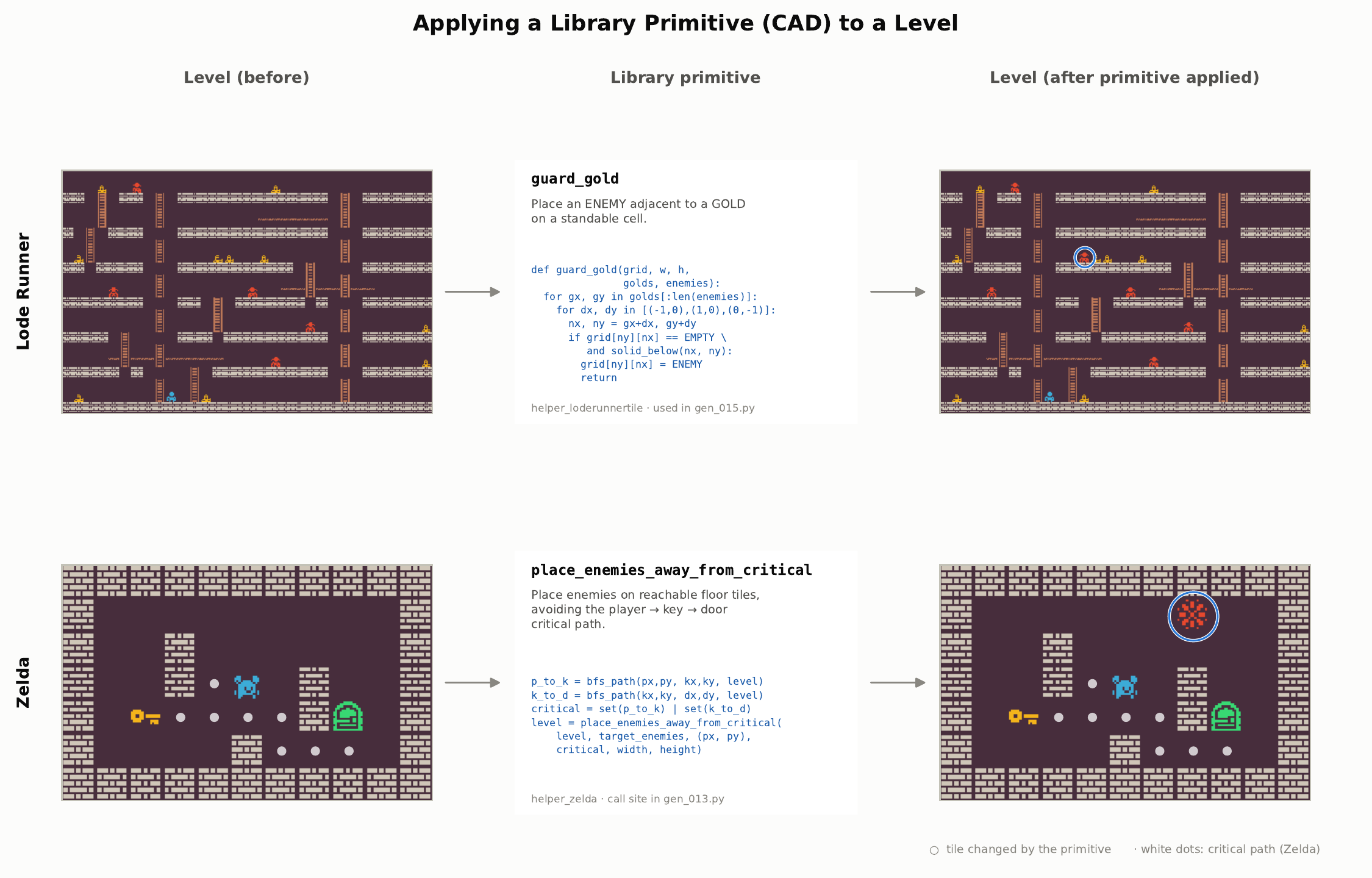}
    \caption{Two primitives from the evolved helper library, each applied to a level (before, left; after, right; changed tile circled): guard\_gold places an enemy beside a gold piece in Lode Runner, and place\_enemies\_away\_from\_critical places an enemy off the player $\rightarrow$ key $\rightarrow$ door critical path (white dots) in Zelda.}
    \label{fig:cad-example}
\end{figure*}

Figure~\ref{fig:cad-example} shows the effect of CAD on the generated levels. For Lode Runner, it abstracted a function that put enemies beside gold to make the levels harder, where the next upcoming generators will make sure to place enemies beside the gold. For Dangerous Dave, something similar happen where CAD added a function that adds enemies/harmful objects around the solution path to make the level harder, which affects the upcoming generators that start adding these objects.

% \begin{table}[t]
% \centering
% \small
% \caption{Complementary Lode Runner warm-start ablation. Values are means $\pm$ standard deviations over ten runs. This experiment is reported separately from the four-domain study.}
% \label{tab:warm-start}
% \resizebox{\columnwidth}{!}{%
% \begin{tabular}{@{}lccc@{}}
% \toprule
% \textbf{Method} & \textbf{Fitness} & \textbf{Solvability} & \textbf{Diversity} \\
% \midrule
% Random Gen. & $0.4393 \pm 0.0718$ & $29.7\% \pm 16.7\%$ & $0.3078 \pm 0.0710$ \\
% Random Gen. + CAD & $0.5019 \pm 0.1179$ & $37.7\% \pm 30.1\%$ & $0.3334 \pm 0.0782$ \\
% Seed Gen. & $0.6131 \pm 0.1170$ & $64.0\% \pm 29.8\%$ & $0.3705 \pm 0.0458$ \\
% Seed Gen. + CAD & $\mathbf{0.6587 \pm 0.0801}$ & $\mathbf{70.7\% \pm 24.6\%}$ & $\mathbf{0.4732 \pm 0.1038}$ \\
% \bottomrule
% \end{tabular}%
% }
% \end{table}

\section{Discussion}

\subsection{Cross-Domain Evidence}

The results show that one evolutionary program-search framework can operate across four games with different grid sizes and benchmark evaluators. CAD improves the mean fitness in every domain. The largest effect appears in Lode Runner when the expert API is available. Search already has domain-specific callable routines in this condition, yet the run-specific learned vocabulary still improves the best fitness. The three other API cells also favor CAD, although their tests do not reach $p<0.05$. 

\subsection{CAD as Representation Adaptation}

CAD changes the operations available to later variation calls during search. The executable substrate remains Python, while the effective vocabulary grows as helpers are extracted. The mechanism measurements show that the resulting functions are adopted by later programs and called repeatedly. Commonly recurring helper functions address patterns such as entity normalization, player placement, reachability, connectivity, and structural construction. These are central operations in tile-based generator design. The large run-specific tail also indicates that CAD does not converge to one fixed library. It combines a recurring core with utilities shaped by the domain and search history.

The program-size analysis supports a related interpretation. CAD shortens the best Base programs, while the fixed API already yields compact source libraries. This does not establish that shorter programs are intrinsically better. It shows that reusable callable vocabularies can offload logic out of the main generator, which may make later edits easier to express and preserve.

\subsection{Practical Implications and Limitations}

Solvability, benchmark quality, and diversity do not fully capture visual style, pacing, novelty, or designer intent. Human review remains important when the desired generator has aesthetic or experiential goals that are absent from the benchmark.

The compute cost is substantial. A typical 50-generation run uses millions of tokens of a near-frontier-class model and several hours of wall-clock time. CAD introduces additional extraction, correction, and refactoring calls. Future work should evaluate fitness gains against token use and wall-clock cost.

CAD is evaluated as a complete pipeline. Helper extraction, module correction, and source refactoring are not separated into individual ablations. The mechanism traces show growth, adoption, and program-size changes, although they do not isolate the causal contribution of each component. Additional work should test longer budgets, other language models, non-tile content, cross-run library transfer, and designer editing of discovered helpers.

\section{Conclusion}

We presented an evolutionary program-search system that uses an LLM to generate executable procedural content generators. The evaluation covers Sokoban, Zelda, Dangerous Dave, and Lode Runner, where each one is run 10 times. Continual Abstraction Discovery raises the mean final best fitness regardless you start with an empty helper library or expert helper functions.

Across the CAD runs, learned libraries grow, are adopted by later programs, and repeatedly recover validation, reachability, and structural utilities. The results support CAD as a practical mechanism for adapting the searchable program vocabulary during procedural content metageneration.

\bibliographystyle{IEEEtran}
\bibliography{bibliography}

\appendix
% \section{Balanced Cross-Domain Trajectory}

% \begin{figure}[t]
%     \centering
%     \includegraphics[width=\columnwidth]{figures/fig_cross_domain_balanced.pdf}
%     \caption{Balanced descriptive summary using ten completed runs per domain. Dashed lines correspond to CAD and solid lines correspond to no CAD. This plot is not used for the all-run Mann Whitney tests in Table~\ref{tab:cad-main}. The internal legend uses the run-directory labels.}
%     \label{fig:balanced-curves}
% \end{figure}

% \section{Generated Level Samples}

% \section{Prompt Templates}

The following figures document the LLM call sites used by the evolutionary loop. Template variables such as \texttt{\{domain\_block\}}, \texttt{\{eval\_feedback\}}, and \texttt{\{parent\_program\}} are filled at runtime.

\begin{figure*}[!t]
    \centering
    \includegraphics[width=0.88\textwidth,height=0.72\textheight,keepaspectratio]{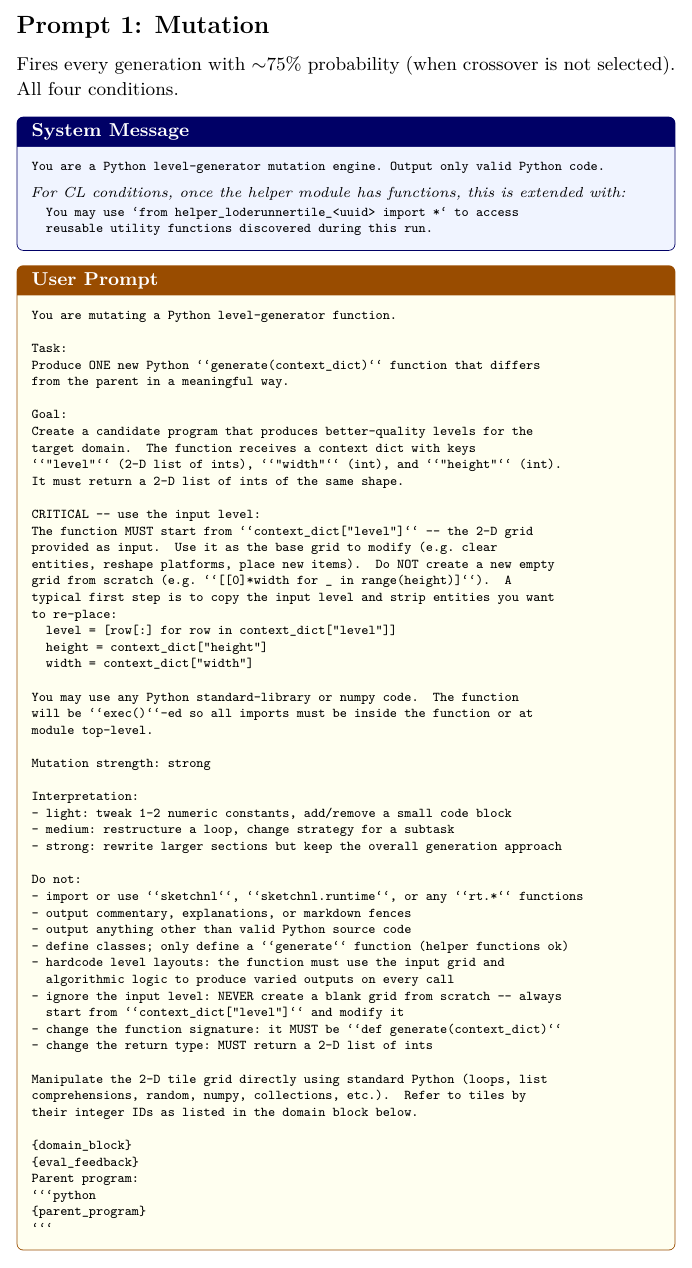}
    \caption{Mutation prompt. The prompt specifies the \texttt{generate(context\_dict)} contract, domain constraints, run memory, available helper APIs, mutation strength, and the parent program.}
    \label{fig:prompt-mutation}
\end{figure*}

\begin{figure*}[!t]
    \centering
    \includegraphics[width=0.88\textwidth,height=0.72\textheight,keepaspectratio]{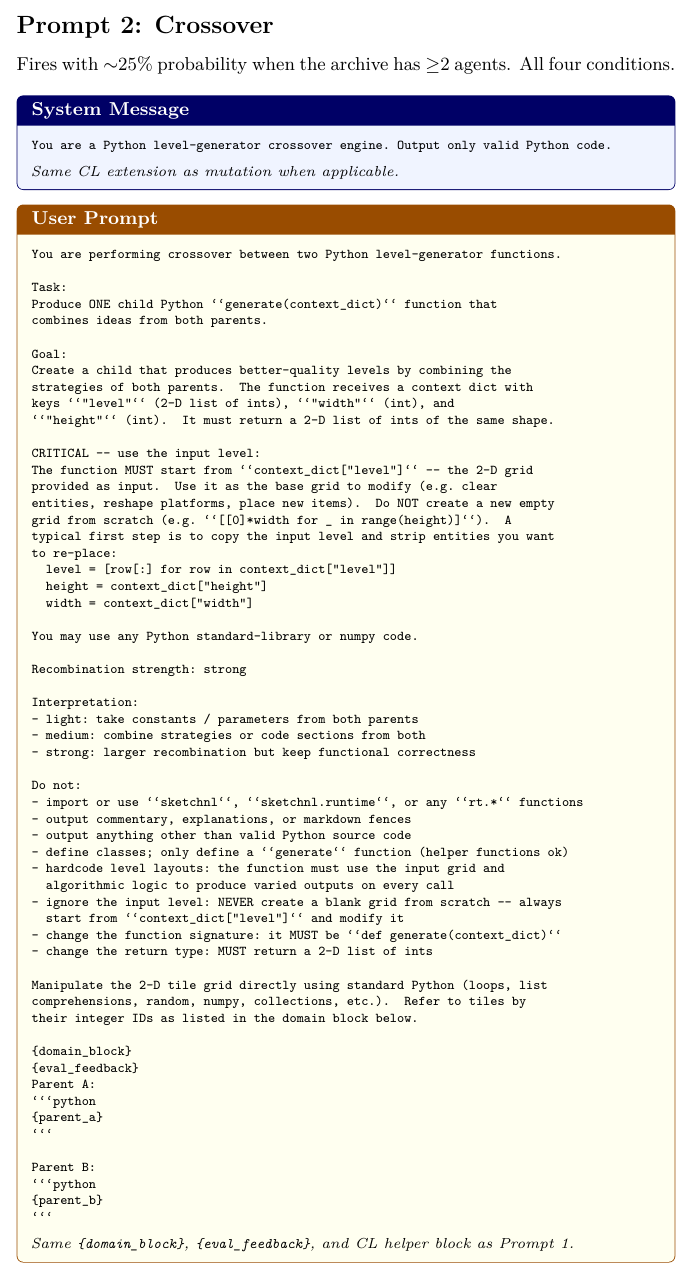}
    \caption{Crossover prompt. The prompt receives two parent programs and asks the model to combine compatible mechanisms into a complete child program.}
    \label{fig:prompt-crossover}
\end{figure*}

\begin{figure*}[!t]
    \centering
    \includegraphics[width=0.88\textwidth,height=0.72\textheight,keepaspectratio]{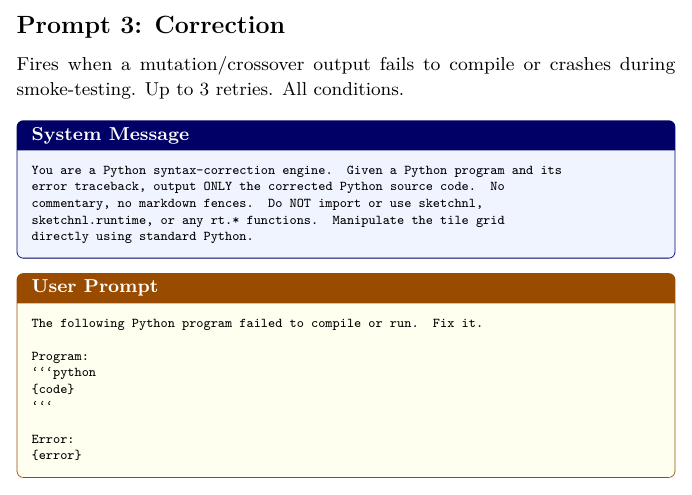}
    \caption{Correction prompt. The prompt is used when a mutation or crossover output fails to compile or crashes during smoke testing. It receives the broken code and the error traceback.}
    \label{fig:prompt-correction}
\end{figure*}

\begin{figure*}[!t]
    \centering
    \includegraphics[width=0.88\textwidth,height=0.72\textheight,keepaspectratio]{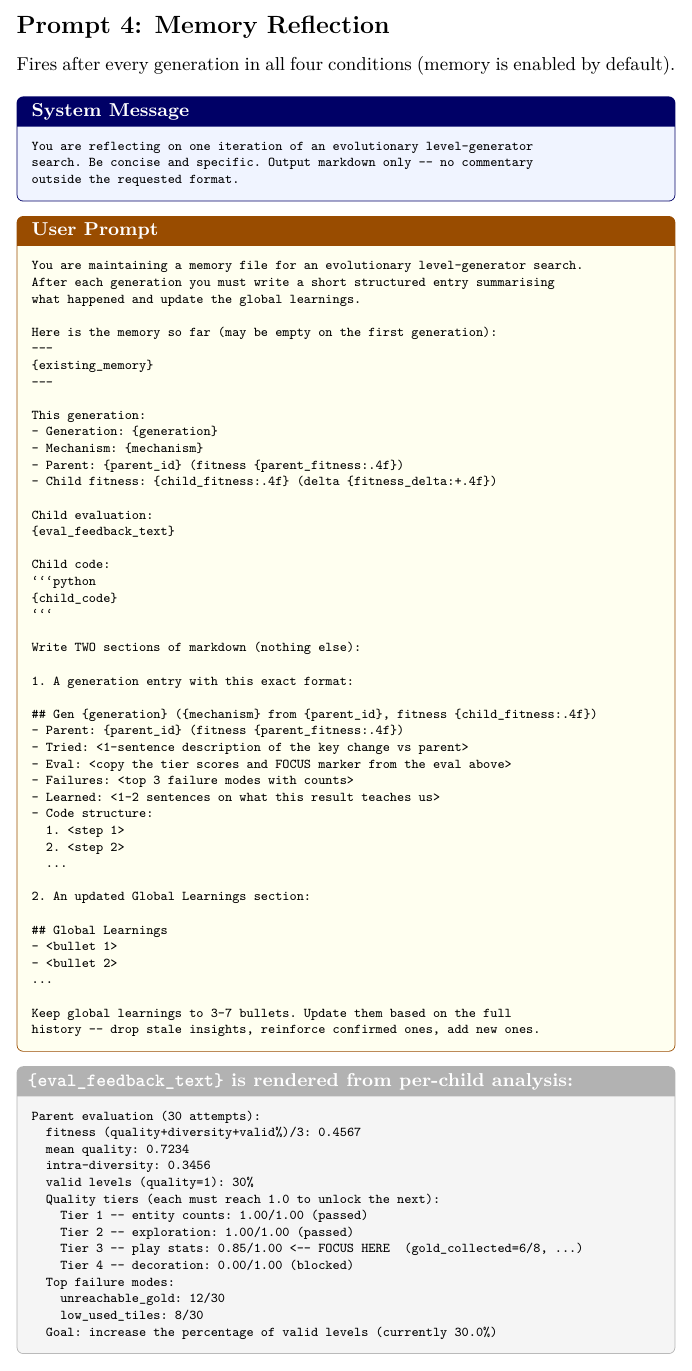}
    \caption{Memory reflection prompt. The prompt maintains a structured per-run memory with generation observations and global lessons for later variation calls.}
    \label{fig:prompt-reflection}
\end{figure*}

\begin{figure*}[!t]
    \centering
    \includegraphics[width=0.88\textwidth,height=0.72\textheight,keepaspectratio]{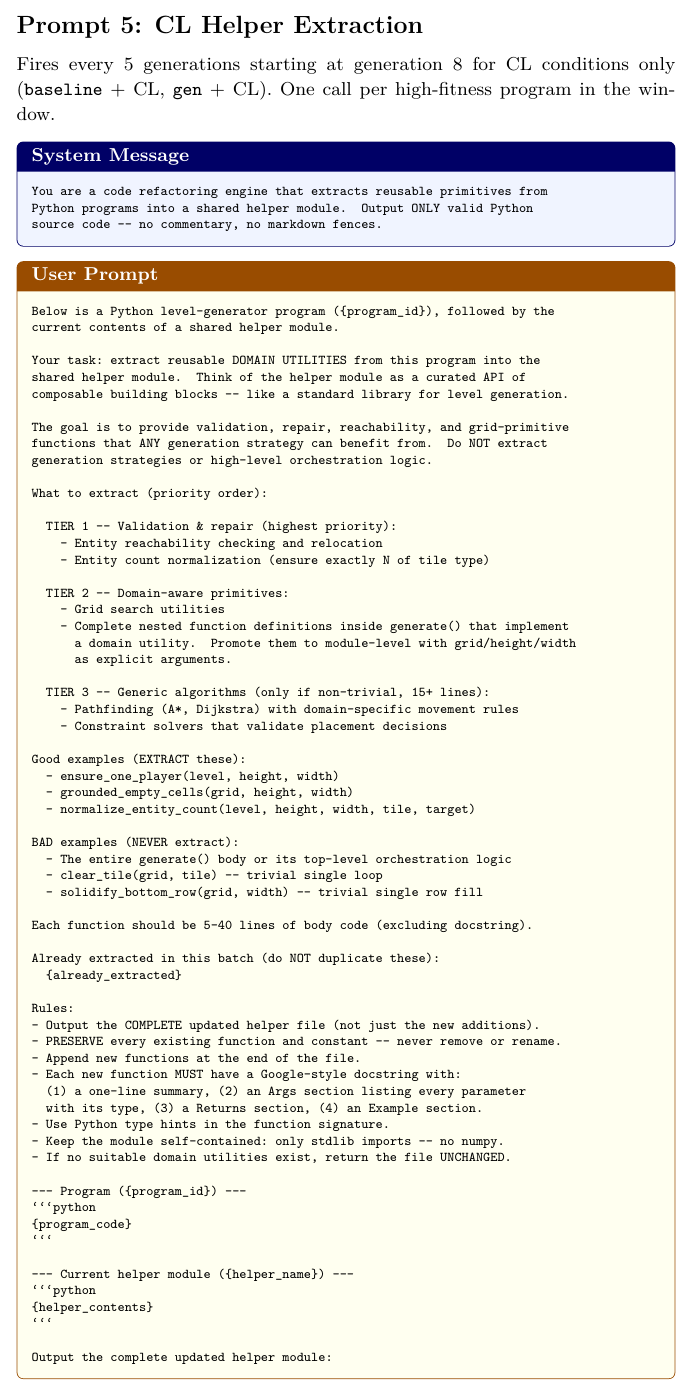}
    \caption{CAD helper extraction prompt. Starting at generation 8 and every five generations thereafter, the prompt extracts reusable domain utilities from programs in the top fitness quartile.}
    \label{fig:prompt-cad-extraction}
\end{figure*}

\begin{figure*}[!t]
    \centering
    \includegraphics[width=0.88\textwidth,height=0.72\textheight,keepaspectratio]{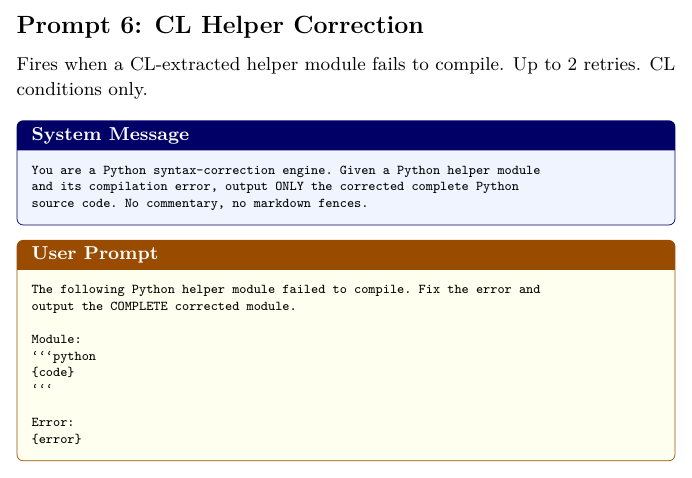}
    \caption{CAD helper correction prompt. The prompt receives the helper module and its compile or smoke-test error.}
    \label{fig:prompt-cad-correction}
\end{figure*}

\begin{figure*}[!t]
    \centering
    \includegraphics[width=0.88\textwidth,height=0.72\textheight,keepaspectratio]{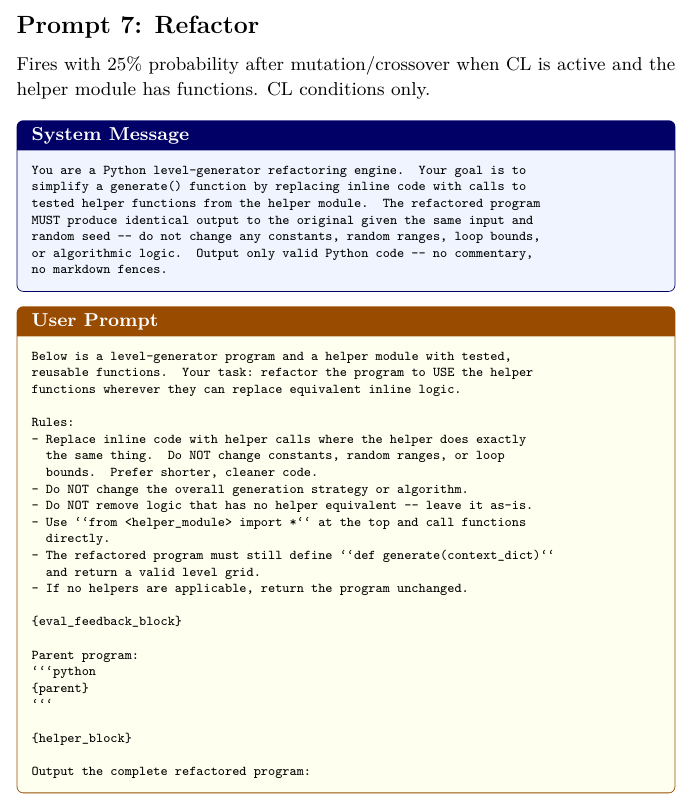}
    \caption{Refactoring prompt. The prompt rewrites a generator to replace duplicated inline logic with calls to tested helper functions.}
    \label{fig:prompt-refactor}
\end{figure*}

\begin{figure*}[!t]
    \centering
    \includegraphics[width=0.88\textwidth,height=0.72\textheight,keepaspectratio]{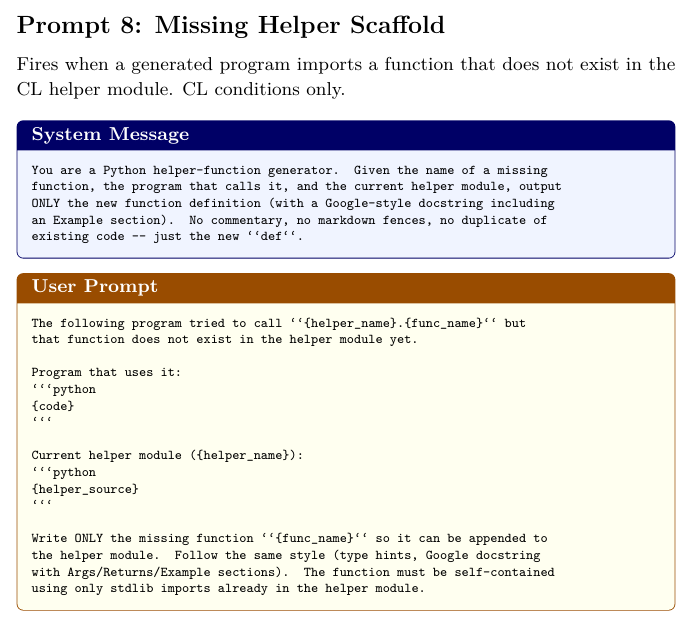}
    \caption{Missing-helper scaffold prompt. The prompt is used when a CAD program imports a helper that does not yet exist in its run-specific module.}
    \label{fig:prompt-missing-helper}
\end{figure*}

\end{document}